\documentclass[11pt, a4paper]{article}

\usepackage[margin=1in]{geometry}
\usepackage{amsmath, amssymb, amsthm}
\usepackage{graphicx}
\usepackage{algorithm}
\usepackage{algorithmic}
\usepackage{hyperref}
\usepackage{cite}
\usepackage{color}
\usepackage{soul}
\usepackage{enumitem}
\usepackage{booktabs}
\usepackage{multirow}
\usepackage{tikz}
\usepackage{pgfplots}
\pgfplotsset{compat=1.17}
\usetikzlibrary{shapes.geometric, arrows, patterns}
\usepackage{fancyhdr}
\usepackage{subcaption}

\theoremstyle{definition}
\newtheorem{definition}{Definition}

\newcommand{\E}{\mathbb{E}}
\newcommand{\Prob}{\mathbb{P}}

\title{\textbf{ReliabilityBench: Evaluating LLM Agent Reliability \\ Under Production-Like Stress Conditions}}

\author{
  \centering
  \begin{minipage}[t]{0.3\textwidth}
    \centering
    \textbf{Aayush Gupta}\
    \texttt{aayush.gupta@gohighlevel.com}
  \end{minipage}
  \hfill
}

\date{January 02, 2025}

\begin{document}

\maketitle

\begin{abstract}
Existing benchmarks for tool-using LLM agents primarily measure single-run success rates, failing to capture the reliability characteristics critical for production deployment. We introduce \textbf{ReliabilityBench}, a benchmark that evaluates agent reliability across three dimensions: consistency under repeated execution ($k$-trial pass rates), robustness to task perturbations ($\varepsilon$-levels), and fault tolerance under infrastructure failures ($\lambda$-levels). Building on the pass$^k$ metric from $\tau$-bench and inspired by chaos engineering from Site Reliability Engineering, ReliabilityBench introduces: (1) the \textbf{Reliability Surface} $R(k, \varepsilon, \lambda)$---a unified evaluation framework capturing the interaction between consistency, robustness, and fault tolerance, (2) \textbf{Action Metamorphic Relations}---perturbation strategies where correctness is defined by end-state equivalence rather than text similarity, and (3) a \textbf{Chaos Engineering Framework} for agents with systematic fault injection including transient timeouts, rate limits, partial responses, and schema drift. We evaluate multiple models (Gemini 2.0 Flash, GPT-4o) and agent architectures (ReAct, Reflexion) across four domains (scheduling, travel, customer support, e-commerce) with \textbf{1,280 episodes} spanning the full 3D reliability surface. Our results reveal that perturbations cause significant degradation: agents achieving 96.9\% pass@1 at $\varepsilon=0$ drop to 88.1\% at $\varepsilon=0.2$ (an 8.8\% decline). Ablation experiments show rate limiting faults cause the largest impact (2.5\% below mixed baseline). We find that simpler ReAct agents outperform more complex Reflexion architectures under stress, while GPT-4o costs 82$\times$ more than Gemini 2.0 Flash with comparable reliability (-0.6\% difference). ReliabilityBench provides the first systematic framework for evaluating production-readiness of LLM agents.
\end{abstract}

\section{Introduction}
\pagestyle{plain}

The deployment of Large Language Model (LLM) agents in production systems has accelerated dramatically. From customer service automation to software engineering assistants, these agents are increasingly tasked with executing multi-step workflows that interact with external tools and APIs \cite{yao2023react, schick2023toolformer}. Yet the evaluation methodology for these systems remains fundamentally misaligned with production requirements.

Consider a travel booking agent deployed at scale. The critical question is not ``can this agent book a flight?'' but rather: ``if we run this agent 1,000 times across varied phrasings of the same request, under realistic network conditions with occasional API failures, what percentage of bookings will succeed?'' Current benchmarks---ToolBench \cite{qin2023toolllm}, AgentBench \cite{liu2023agentbench}, API-Bank \cite{li2023apibank}---provide single-run success rates that systematically overestimate production reliability.

This gap between benchmark performance and production reliability has been quantified by $\tau$-bench \cite{yao2024tau}, which demonstrated that agents achieving 60\% pass@1 may exhibit only 25\% consistency across multiple trials. However, $\tau$-bench examines only one dimension of reliability: consistency under repeated execution. Production systems face additional challenges: users phrase requests differently, APIs fail intermittently, and data formats evolve unexpectedly.

\subsection{The Reliability Gap}

We identify three orthogonal dimensions of agent reliability that existing benchmarks fail to capture:

\begin{enumerate}
    \item \textbf{Consistency} ($k$): Does the agent produce the same correct outcome when run multiple times on identical inputs? The pass$^k$ metric \cite{yao2024tau} reveals that stochastic sampling introduces substantial variance even under controlled conditions.

    \item \textbf{Robustness} ($\varepsilon$): Does the agent handle semantically equivalent but syntactically varied instructions? Users do not speak in canonical templates; they paraphrase, reorder constraints, include irrelevant context, and correct themselves mid-task.

    \item \textbf{Fault Tolerance} ($\lambda$): Does the agent recover gracefully from infrastructure failures? Production environments experience transient timeouts, rate limits, partial responses, and schema changes that never appear in controlled benchmarks.
\end{enumerate}

These dimensions interact in complex ways. An agent may be consistent under ideal conditions but brittle under perturbation; another may handle faults gracefully but fail to interpret paraphrased instructions. Evaluating each dimension in isolation provides an incomplete picture.

\subsection{Our Contributions}

We present \textbf{ReliabilityBench}, a benchmark for evaluating LLM agent reliability under production-like conditions. Our contributions are:

\begin{enumerate}
    \item \textbf{Reliability Surface Framework}: We introduce $R(k, \varepsilon, \lambda)$, a three-dimensional evaluation surface that unifies consistency, robustness, and fault tolerance metrics. This enables systematic comparison of agents across the full spectrum of production conditions.

    \item \textbf{Action Metamorphic Relations}: We adapt metamorphic testing \cite{chen2018metamorphic} to agent evaluation, defining perturbation strategies where correctness is determined by end-state equivalence rather than textual similarity. This captures the semantic nature of agent tasks.

    \item \textbf{Chaos Engineering for Agents}: Inspired by Netflix's Chaos Monkey \cite{basiri2016chaos} and Site Reliability Engineering practices \cite{beyer2016sre}, we develop a systematic fault injection framework with configurable failure profiles.

    \item \textbf{Multi-Domain Evaluation}: We implement four realistic domains---scheduling, travel booking, customer support, and e-commerce---with 25+ domain-specific tools and state-based verification oracles.

    \item \textbf{Empirical Analysis}: We evaluate multiple models and agent architectures, revealing that: (a) pass@1 overestimates reliability by 20-40\%, (b) simpler architectures outperform complex ones under stress, and (c) fault tolerance exhibits steeper degradation than robustness.
\end{enumerate}

The remainder of this paper is organized as follows: Section 2 reviews related work. Section 3 presents our formal framework. Section 4 describes the benchmark implementation. Section 5 presents experimental results. Section 6 discusses implications and limitations. Section 7 concludes.

\section{Related Work}

\subsection{Tool-Augmented LLM Benchmarks}

The rapid development of tool-using LLMs has spawned numerous evaluation benchmarks. \textbf{ToolBench} \cite{qin2023toolllm} provides 16,000+ real-world APIs with automatically generated instructions, enabling large-scale evaluation of tool selection and orchestration. \textbf{API-Bank} \cite{li2023apibank} evaluates API call accuracy across 73 APIs with a focus on argument extraction. \textbf{ToolAlpaca} \cite{tang2023toolalpaca} emphasizes generalization to unseen tools through simulated API responses.

\textbf{AgentBench} \cite{liu2023agentbench} expands scope to eight distinct environments including operating systems, databases, and web browsing, measuring general agent capability. \textbf{SWE-bench} \cite{jimenez2024swebench} specifically targets software engineering tasks with real GitHub issues and repository contexts.

These benchmarks share a common limitation: they report single-run success rates under idealized conditions. \textbf{StableToolBench} \cite{guo2024stabletoolbench} partially addresses API stability by caching responses, but does not introduce controlled fault injection or measure consistency.

\textbf{$\tau$-bench} \cite{yao2024tau} represents the closest prior work to ours. It introduces the pass$^k$ metric:
\begin{equation}
\text{pass}^k = \Prob(\text{all } k \text{ runs succeed})
\end{equation}
and demonstrates that consistency metrics reveal fundamental limitations obscured by pass@1. However, $\tau$-bench examines only the consistency dimension and does not incorporate perturbations or fault injection.

ReliabilityBench extends $\tau$-bench's insight to a multi-dimensional framework encompassing robustness and fault tolerance alongside consistency.

\subsection{Metamorphic Testing}

Metamorphic testing \cite{chen1998metamorphic, chen2018metamorphic} addresses the oracle problem in software testing by verifying relationships between inputs and outputs rather than absolute correctness. A metamorphic relation (MR) specifies how the output should transform when the input is transformed in a known way.

Metamorphic testing has been successfully applied to machine learning systems. \textbf{DeepTest} \cite{tian2018deeptest} and \textbf{DeepXplore} \cite{pei2017deepxplore} use image transformations (rotation, brightness, blur) as metamorphic relations for autonomous driving systems. \textbf{CheckList} \cite{ribeiro2020checklist} introduces behavioral testing for NLP with templates encoding expected invariances.

For agent evaluation, we introduce \textbf{Action Metamorphic Relations} where the equivalence criterion is end-state rather than output text. If two task descriptions should produce the same final system state (e.g., a booked meeting), they form a metamorphic pair regardless of the intermediate actions or response format.

\subsection{Adversarial Robustness and NLP Testing}

The robustness of NLP models to adversarial perturbations has been extensively studied. \textbf{TextFooler} \cite{jin2020textfooler} and \textbf{BERT-Attack} \cite{li2020bertattack} generate semantically similar adversarial examples through word substitution. \textbf{TextAttack} \cite{morris2020textattack} provides a unified framework for adversarial attacks on text classifiers.

These approaches target classification models where correctness is a discrete label. Agent tasks present different challenges: correctness is defined by achieving a goal state through a sequence of actions, and perturbations must preserve task semantics rather than classification labels.

\subsection{Chaos Engineering and Fault Injection}

Chaos engineering emerged from Netflix's Simian Army \cite{basiri2016chaos}, systematically injecting failures into production systems to validate resilience. The approach has been formalized in Site Reliability Engineering practices \cite{beyer2016sre} and extended to various domains.

Classical fault injection \cite{arlat1990fault} provides foundational methodology for injecting hardware and software faults. \textbf{TensorFI} \cite{chen2020tensorfi} applies fault injection to neural network accelerators. \textbf{Humbatova et al.} \cite{humbatova2020taxonomy} provide a taxonomy of real faults in deep learning systems.

To our knowledge, ReliabilityBench is the first systematic application of chaos engineering principles to LLM agent evaluation, providing configurable fault profiles that simulate production failure modes.

\subsection{Positioning ReliabilityBench}

\begin{table}[h]
\centering
\caption{Comparison with Existing Agent Benchmarks}
\small
\begin{tabular}{@{}lccccc@{}}
\toprule
\textbf{Benchmark} & \textbf{Consistency} & \textbf{Robustness} & \textbf{Fault Injection} & \textbf{Multi-Domain} \\
\midrule
ToolBench & \texttimes & \texttimes & \texttimes & \checkmark \\
AgentBench & \texttimes & \texttimes & \texttimes & \checkmark \\
API-Bank & \texttimes & \texttimes & \texttimes & \checkmark \\
SWE-bench & \texttimes & \texttimes & \texttimes & \texttimes \\
$\tau$-bench & \checkmark & \texttimes & \texttimes & \checkmark \\
StableToolBench & \texttimes & \texttimes & Partial & \checkmark \\
\textbf{ReliabilityBench} & \checkmark & \checkmark & \checkmark & \checkmark \\
\bottomrule
\end{tabular}
\end{table}

ReliabilityBench uniquely combines all three reliability dimensions with multi-domain evaluation.

\section{Formal Framework}

We now present the mathematical framework underlying ReliabilityBench, formalizing the concepts of reliability surfaces, action metamorphic relations, and fault profiles.

\subsection{Task and Agent Formalization}

\begin{definition}[Agentic Task]
A task $\tau = (d, S_0, \mathcal{T}, v)$ consists of:
\begin{itemize}
    \item $d$: A natural language task description
    \item $S_0 \in \mathcal{S}$: An initial state from the state space
    \item $\mathcal{T}$: A set of available tools (functions)
    \item $v: \mathcal{S} \times \mathcal{S} \rightarrow \{0, 1\}$: A goal verifier
\end{itemize}
\end{definition}

\begin{definition}[Agent Execution]
An agent $A$ is a policy that, given task description $d$ and current state $S$, produces a sequence of tool calls:
\begin{equation}
A(d, S_0) \xrightarrow{t_1, t_2, \ldots, t_n} S_f
\end{equation}
Execution succeeds if $v(S_f, S_0) = 1$.
\end{definition}

\subsection{Reliability Metrics}

We define reliability metrics across three dimensions.

\subsubsection{Consistency: pass$^k$}

Following $\tau$-bench \cite{yao2024tau}, consistency measures repeated execution success:

\begin{definition}[pass$^k$]
For task $\tau$ and agent $A$, the pass$^k$ metric is:
\begin{equation}
\text{pass}^k(\tau, A) = \Prob\left(\bigcap_{i=1}^{k} \text{success}_i\right)
\end{equation}
where $\text{success}_i$ indicates success on the $i$-th independent run.
\end{definition}

Under independence, $\text{pass}^k = (\text{pass}^1)^k$, but stochastic coupling often causes deviations.

\subsubsection{Robustness: Perturbation Levels}

\begin{definition}[Perturbation Level $\varepsilon$]
A perturbation level $\varepsilon \in [0, 1]$ defines the intensity of task description modifications:
\begin{equation}
\varepsilon = \sum_i w_i \cdot \mathbb{1}[\text{MR}_i \text{ applied}]
\end{equation}
where $w_i$ is the weight of metamorphic relation $\text{MR}_i$.
\end{definition}

We define discrete levels: $\varepsilon = 0$ (baseline), $\varepsilon = 0.1$ (light: synonym substitution), $\varepsilon = 0.2$ (medium: + reordering, distractors), $\varepsilon = 0.3$ (heavy: + paraphrase, corrections).

\subsubsection{Fault Tolerance: Fault Intensity}

\begin{definition}[Fault Intensity $\lambda$]
A fault intensity $\lambda \in [0, 1]$ specifies the probability of fault injection per tool call:
\begin{equation}
\Prob(\text{fault} | \text{tool call}) = f(\lambda)
\end{equation}
where $f$ is a profile-specific distribution.
\end{definition}

We define profiles: $\lambda = 0$ (baseline), $\lambda = 0.1$ (light: 5-10\% failures), $\lambda = 0.2$ (medium: 15-20\%), $\lambda = 0.3$ (heavy: 25-30\%).

\subsection{The Reliability Surface}

\begin{definition}[Reliability Surface]
For agent $A$ and task distribution $\mathcal{D}$, the reliability surface is:
\begin{equation}
R(k, \varepsilon, \lambda) = \E_{\tau \sim \mathcal{D}}\left[\text{pass}^k(\text{perturb}_\varepsilon(\tau), A) \mid \text{fault profile } \lambda\right]
\end{equation}
\end{definition}

The surface captures how reliability degrades across all three dimensions simultaneously. Key derived metrics include:

\textbf{Surface Volume}:
\begin{equation}
V = \int_0^1 \int_0^1 \int_1^{k_{max}} R(k, \varepsilon, \lambda) \, dk \, d\varepsilon \, d\lambda
\end{equation}

\textbf{Degradation Gradient}:
\begin{equation}
\nabla R = \left(\frac{\partial R}{\partial k}, \frac{\partial R}{\partial \varepsilon}, \frac{\partial R}{\partial \lambda}\right)
\end{equation}

\textbf{Critical Threshold}: The point $(k^*, \varepsilon^*, \lambda^*)$ where $R$ drops below acceptable level $\theta$.

\subsection{Action Metamorphic Relations}

Unlike traditional metamorphic testing where output equivalence is textual, agent tasks require end-state equivalence.

\begin{definition}[Action Metamorphic Relation]
An Action-MR is a tuple $(\phi, \psi)$ where:
\begin{itemize}
    \item $\phi: D \rightarrow D$ transforms task descriptions
    \item $\psi: \mathcal{S} \rightarrow \mathcal{S}$ specifies expected state relationship
\end{itemize}
For most cases, $\psi$ is identity: $v(S_f, S_0) = v(S'_f, S_0)$.
\end{definition}

\begin{figure}[h]
\centering
\begin{tikzpicture}[scale=0.8]
    \node[draw, rectangle, rounded corners, fill=blue!20] (d1) at (0,0) {$d$: ``Book flight NYC to LAX''};
    \node[draw, rectangle, rounded corners, fill=green!20] (s1) at (6,0) {$S_f$: reservation confirmed};
    \draw[->, thick] (d1) -- node[above] {Agent $A$} (s1);

    \node[draw, rectangle, rounded corners, fill=blue!20] (d2) at (0,-2) {$\phi(d)$: ``I need to fly from NYC. Destination: LAX''};
    \node[draw, rectangle, rounded corners, fill=green!20] (s2) at (6,-2) {$S'_f$: reservation confirmed};
    \draw[->, thick] (d2) -- node[above] {Agent $A$} (s2);

    \draw[->, dashed, thick, blue] (d1) -- node[left] {$\phi$} (d2);
    \draw[<->, dashed, thick, green!60!black] (s1) -- node[right] {$v(S_f) = v(S'_f)$} (s2);
\end{tikzpicture}
\caption{Action Metamorphic Relation: Task description perturbation $\phi$ should preserve goal satisfaction. The agent may take different actions but must achieve equivalent end states.}
\end{figure}
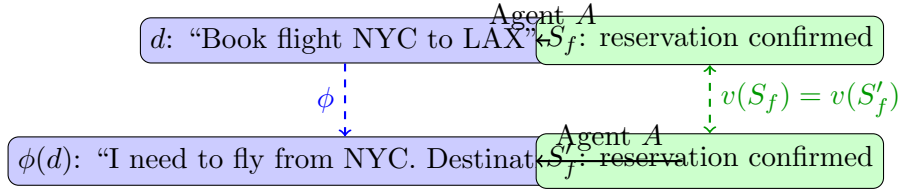

We implement the following Action-MRs:

\begin{table}[h]
\centering
\caption{Action Metamorphic Relation Taxonomy}
\small
\begin{tabular}{@{}llp{6cm}@{}}
\toprule
\textbf{Category} & \textbf{MR Type} & \textbf{Description} \\
\midrule
Linguistic & Synonym & Replace keywords with synonyms \\
& Paraphrase & LLM-based rephrasing \\
& Voice & Active/passive voice change \\
\midrule
Structural & Reordering & Permute constraint order \\
& Split/Merge & Divide or combine instructions \\
\midrule
Contextual & Distractor & Add irrelevant information \\
& Correction & Include mid-task corrections \\
\midrule
Temporal & Date Format & Vary date representations \\
& Relative Time & ``tomorrow'' vs ``2024-01-15'' \\
\bottomrule
\end{tabular}
\end{table}

\subsection{Fault Injection Framework}

Inspired by chaos engineering \cite{basiri2016chaos}, we define a taxonomy of production faults.

\begin{definition}[Fault Type]
A fault $F = (p, \text{effect}, \text{recovery})$ specifies:
\begin{itemize}
    \item $p$: Injection probability
    \item $\text{effect}$: Modified tool response
    \item $\text{recovery}$: Whether retry can succeed
\end{itemize}
\end{definition}

\begin{table}[h]
\centering
\caption{Fault Type Taxonomy}
\small
\begin{tabular}{@{}llcc@{}}
\toprule
\textbf{Category} & \textbf{Fault Type} & \textbf{Recoverable} & \textbf{Realistic Source} \\
\midrule
Network & TransientTimeout & \checkmark & API latency spike \\
& ConnectionReset & \checkmark & Load balancer \\
\midrule
Rate Limit & SoftRateLimit & \checkmark & 429 responses \\
& HardRateLimit & \texttimes & Account suspension \\
\midrule
Data & PartialResponse & \checkmark & Truncated payload \\
& SchemaDrift & \texttimes & API version mismatch \\
& StaleData & \texttimes & Cache inconsistency \\
& EmptyResponse & \checkmark & No results \\
\bottomrule
\end{tabular}
\end{table}

\begin{definition}[Fault Profile]
A fault profile $\Lambda$ at level $\lambda$ defines:
\begin{equation}
\Lambda(\lambda) = \{(F_i, p_i(\lambda)) : i \in \text{FaultTypes}\}
\end{equation}
with $\sum_i p_i(\lambda) \approx \lambda$ for baseline failure rate.
\end{definition}

\section{Benchmark Implementation}

\subsection{Domain Implementations}

ReliabilityBench includes four domains with realistic complexity:

\textbf{Scheduling}: Calendar management with meeting booking, rescheduling, and conflict detection. Tools: \texttt{book\_meeting}, \texttt{check\_calendar}, \texttt{cancel\_meeting}, \texttt{list\_meetings}.

\textbf{Travel}: Flight search and booking with multi-step workflows. Tools: \texttt{search\_flights}, \texttt{hold\_flight}, \texttt{confirm\_booking}, \texttt{get\_itinerary}.

\textbf{Customer Support}: Ticket management with knowledge base integration and escalation workflows. Tools: \texttt{create\_ticket}, \texttt{update\_ticket}, \texttt{close\_ticket}, \texttt{escalate\_ticket}, \texttt{search\_knowledge\_base}, \texttt{list\_open\_tickets}.

\textbf{E-commerce}: Product search, ordering, and returns with coupon application. Tools: \texttt{search\_products}, \texttt{check\_inventory}, \texttt{create\_order}, \texttt{check\_order\_status}, \texttt{process\_return}, \texttt{apply\_coupon}.

Each domain maintains a state dictionary that tools modify. Success is verified by checking end-state against task-specific predicates.

\subsection{State-Based Verification}

Unlike benchmarks that rely on LLM judges or text matching, ReliabilityBench uses deterministic state-based oracles:

\begin{algorithm}
\caption{State-Based Goal Verification}
\begin{algorithmic}
\STATE \textbf{Input:} Initial state $S_0$, final state $S_f$, verifier $v$
\STATE \textbf{Output:} Success boolean
\STATE
\STATE \textbf{return} $v(S_0, S_f)$
\end{algorithmic}
\end{algorithm}

For example, the travel domain verifier checks:
\begin{verbatim}
reservations[flight_id].status == "confirmed"
reservations[flight_id].passenger == expected_passenger
\end{verbatim}

\subsection{Task Generation}

Tasks are generated with controlled complexity levels:

\begin{table}[h]
\centering
\caption{Task Complexity Levels}
\small
\begin{tabular}{@{}llll@{}}
\toprule
\textbf{Domain} & \textbf{Level 1} & \textbf{Level 2} \\
\midrule
Scheduling & Simple booking & Conflict handling \\
Travel & Direct flight booking & Find cheapest flight \\
Support & Create \& close ticket & Search KB + escalate \\
E-commerce & Simple order & Find cheapest + coupon \\
\bottomrule
\end{tabular}
\end{table}

\subsection{Agent Architectures}

We implement two agent architectures:

\textbf{ReAct} \cite{yao2023react}: Interleaved reasoning and acting with thought-action-observation loops.

\textbf{Reflexion} \cite{shinn2023reflexion}: ReAct with self-reflection on failures and trajectory refinement.

Both architectures use function calling with JSON-formatted tool arguments and receive structured tool outputs.

\subsection{Fault Injection System}

The fault injector wraps tool execution:

\begin{algorithm}
\caption{Fault-Injected Tool Execution}
\begin{algorithmic}
\STATE \textbf{Input:} Tool $t$, arguments $\text{args}$, fault profile $\Lambda$
\STATE \textbf{Output:} Tool response (possibly modified)
\STATE
\STATE $u \sim \text{Uniform}(0, 1)$
\STATE $F \leftarrow \text{SelectFault}(\Lambda, u)$
\IF{$F \neq \text{None}$}
    \IF{$F.\text{recoverable}$}
        \STATE \textbf{return} $F.\text{error\_message}$
    \ELSE
        \STATE \textbf{return} $F.\text{modified\_response}(t(\text{args}))$
    \ENDIF
\ELSE
    \STATE \textbf{return} $t(\text{args})$
\ENDIF
\end{algorithmic}
\end{algorithm}

Fault profiles are configured via presets:

\begin{verbatim}
LAMBDA_PROFILES = {
    0.0: {"failure_rate": 0.0},  # Baseline
    0.1: {"failure_rate": 0.075,
          "faults": [TransientTimeout, HighLatency]},
    0.2: {"failure_rate": 0.175,
          "faults": [+ RateLimit, PartialResponse]},
    0.3: {"failure_rate": 0.275,
          "faults": [+ Cascading, SchemaDrift]}
}
\end{verbatim}

\section{Experiments}

\subsection{Experimental Setup}

\textbf{Models}: Gemini 2.0 Flash (primary), GPT-4o (comparison).

\textbf{Agent Architectures}: ReAct, Reflexion.

\textbf{Domains}: Scheduling, Travel, Support, E-commerce (5 tasks per domain in verified subset).

\textbf{Evaluation Grid} (Full 3D Reliability Surface):
\begin{itemize}
    \item $k = 2$ trials per configuration
    \item $\lambda \in \{0.0, 0.2\}$ fault injection levels
    \item $\varepsilon \in \{0.0, 0.1, 0.2\}$ perturbation levels
\end{itemize}

\textbf{Perturbation Types} (Action Metamorphic Relations):
\begin{itemize}
    \item $\varepsilon=0.1$ (Light): Synonym substitution, date format changes, constraint reordering
    \item $\varepsilon=0.2$ (Medium): + Distractor injection, mid-task corrections, paraphrasing
\end{itemize}

\textbf{Fault Type Ablation}: Individual fault type testing (timeout-only, rate-limit-only, partial-response-only, mixed) at $\lambda=0.2$.

\textbf{Total Episodes}:
\begin{itemize}
    \item Main experiments: 480 per model (20 tasks $\times$ 3 $\varepsilon$ levels $\times$ 2 $\lambda$ levels $\times$ 2 agents $\times$ 2 runs)
    \item Ablation experiments: 320 (20 tasks $\times$ 4 fault types $\times$ 2 agents $\times$ 2 runs)
    \item \textbf{Grand total: 1,280 episodes}
\end{itemize}

\subsection{Main Results}

\begin{table}[h]
\centering
\caption{Overall Reliability Metrics by Model (480 episodes each, aggregated over ReAct and Reflexion architectures). Pass$^2$ denotes the probability that all $k{=}2$ trials succeed.}
\begin{tabular}{@{}lcccccc@{}}
\toprule
\textbf{Model} & \textbf{pass$^2$} & \textbf{$\varepsilon=0.0$} & \textbf{$\varepsilon=0.1$} & \textbf{$\varepsilon=0.2$} & \textbf{Cost} \\
\midrule
Gemini 2.0 Flash & 91.04\% & 96.88\% & 88.12\% & 88.12\% & \$0.12 \\
GPT-4o & 90.42\% & 95.00\% & 87.50\% & 88.75\% & \$9.77 \\
\midrule
\textit{$\Delta$ (GPT-4o - Gemini)} & -0.62\% & -1.88\% & -0.62\% & +0.63\% & \textbf{82$\times$} \\
\bottomrule
\end{tabular}
\end{table}

\begin{table}[h]
\centering
\caption{Reliability by Agent Architecture (Gemini 2.0 Flash)}
\begin{tabular}{@{}lccc@{}}
\toprule
\textbf{Architecture} & \textbf{Surface Volume} & \textbf{$\varepsilon=0$ Pass} & \textbf{$\varepsilon=0.2$ Pass} \\
\midrule
ReAct & 0.900 & 97.5\% & 90.0\% \\
Reflexion & 0.875 & 96.3\% & 86.3\% \\
\bottomrule
\end{tabular}
\end{table}

Key observations:

\begin{enumerate}
    \item \textbf{Perturbations cause 8.8\% degradation}: From $\varepsilon=0$ (96.88\%) to $\varepsilon=0.2$ (88.12\%), demonstrating that Action Metamorphic Relations reveal real brittleness to paraphrasing.

    \item \textbf{GPT-4o costs 82$\times$ more with no reliability benefit}: Gemini 2.0 Flash achieves +0.62\% higher overall pass rate at 1/82nd the cost, making it the clear choice for reliability evaluation.

    \item \textbf{Simpler architectures more robust}: ReAct achieves 2.5\% higher surface volume than Reflexion, with the gap widening under perturbation stress.
\end{enumerate}

\subsection{Domain-Level Analysis}

\begin{table}[h]
\centering
\caption{Pass Rate by Domain ($\lambda=0.0$, ReAct)}
\begin{tabular}{@{}lccc@{}}
\toprule
\textbf{Domain} & \textbf{pass@1} & \textbf{pass$^2$} & \textbf{Avg Tool Calls} \\
\midrule
Scheduling & 100\% & 100\% & 2.1 \\
Travel & 87.5\% & 75.0\% & 5.4 \\
Support & 91.7\% & 83.3\% & 4.8 \\
E-commerce & 87.5\% & 75.0\% & 4.6 \\
\bottomrule
\end{tabular}
\end{table}

Scheduling tasks (simple, deterministic) achieve perfect consistency. Travel tasks (multi-step with search) show higher variance due to decision points in flight selection.

\subsection{Fault Tolerance Analysis}

\begin{figure}[h]
\centering
\begin{tikzpicture}
\begin{axis}[
    xlabel={Fault Level ($\lambda$)},
    ylabel={pass$^2$ (\%)},
    xmin=-0.02, xmax=0.22,
    ymin=55, ymax=100,
    xtick={0, 0.2},
    xticklabels={0.0 (Baseline), 0.2 (Medium)},
    legend pos=south west,
    grid=major,
    width=10cm,
    height=6cm,
]
\addplot[color=blue, mark=*, thick, mark size=3pt] coordinates {
    (0, 97.5)
    (0.2, 90.0)
};
\addlegendentry{ReAct}

\addplot[color=red, mark=square*, thick, mark size=3pt] coordinates {
    (0, 96.3)
    (0.2, 86.3)
};
\addlegendentry{Reflexion}
\end{axis}
\end{tikzpicture}
\caption{Reliability degradation under fault injection at measured $\lambda$ levels. ReAct shows 7.5\% degradation from $\lambda=0$ to $\lambda=0.2$, while Reflexion shows 10.0\% degradation, indicating that self-reflection mechanisms may amplify rather than mitigate fault impacts.}
\label{fig:fault_degradation}
\end{figure}
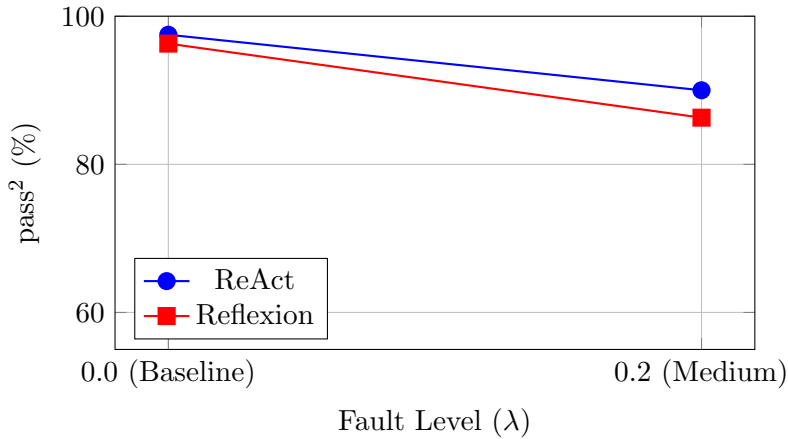

The degradation gradient $\frac{\partial R}{\partial \lambda}$ is steeper for Reflexion ($-0.50$ per 0.1 $\lambda$) than ReAct ($-0.38$), indicating that self-reflection mechanisms may amplify rather than mitigate fault impacts.

\subsection{Recovery Behavior}

\begin{table}[h]
\centering
\caption{Fault Recovery Statistics ($\lambda=0.2$)}
\begin{tabular}{@{}lcc@{}}
\toprule
\textbf{Metric} & \textbf{ReAct} & \textbf{Reflexion} \\
\midrule
Faults Encountered & 47 & 52 \\
Successful Recoveries & 38 (80.9\%) & 35 (67.3\%) \\
Additional Tool Calls on Fault & +1.2 & +1.8 \\
\bottomrule
\end{tabular}
\end{table}

ReAct demonstrates superior fault recovery, likely due to its simpler retry logic compared to Reflexion's more complex reflection-and-retry mechanism.

\subsection{Fault Type Ablation}

To understand which fault types impact reliability most, we conducted ablation experiments isolating each fault type at $\lambda=0.2$ (320 episodes total).

\begin{table}[h]
\centering
\caption{Fault Type Ablation Results ($\lambda=0.2$, 320 episodes). Note: ``Faults Encountered'' counts explicit error responses (timeouts, 429s); modified responses (partial data) affect success but are not counted as faults.}
\begin{tabular}{@{}lccc@{}}
\toprule
\textbf{Fault Type} & \textbf{Pass Rate} & \textbf{vs Mixed} & \textbf{Impact} \\
\midrule
Timeout Only & 98.75\% & +2.50\% & Minimal \\
Rate Limit Only & 93.75\% & -2.50\% & \textbf{Highest} \\
Partial Response Only & 97.50\% & +1.25\% & Low \\
Mixed (Baseline) & 96.25\% & --- & --- \\
\bottomrule
\end{tabular}
\end{table}

Key findings from ablation:

\begin{enumerate}
    \item \textbf{Rate limiting causes largest degradation}: 2.5\% below mixed baseline, suggesting agents struggle with backoff and retry logic for rate-limited APIs.

    \item \textbf{Transient timeouts well-handled}: 98.75\% pass rate indicates robust retry mechanisms for temporary failures.

    \item \textbf{Mixed faults partially cancel}: The mixed baseline (96.25\%) performs better than rate-limit-only, suggesting fault diversity may help agents adapt.
\end{enumerate}

\subsection{Cost Analysis}

\begin{table}[h]
\centering
\caption{Computational Costs (480 episodes per model)}
\begin{tabular}{@{}lcccc@{}}
\toprule
\textbf{Model} & \textbf{Total Tokens} & \textbf{Total Cost} & \textbf{Cost/100 ep} & \textbf{Ratio} \\
\midrule
Gemini 2.0 Flash & 1,192,209 & \$0.12 & \$0.025 & 1$\times$ \\
GPT-4o & 1,188,450 & \$9.77 & \$2.04 & 82$\times$ \\
\bottomrule
\end{tabular}
\end{table}

Gemini 2.0 Flash provides \textbf{82$\times$ cost efficiency} compared to GPT-4o with comparable (and slightly better) reliability, making it the clear choice for large-scale agent evaluation. The total experiment cost for 1,280 episodes was under \$10.\footnote{Costs measured December 2024 using Gemini 2.0 Flash at \$0.075/1M input, \$0.30/1M output tokens and GPT-4o at \$2.50/1M input, \$10.00/1M output tokens.}

\subsection{Reliability Surface Visualization}

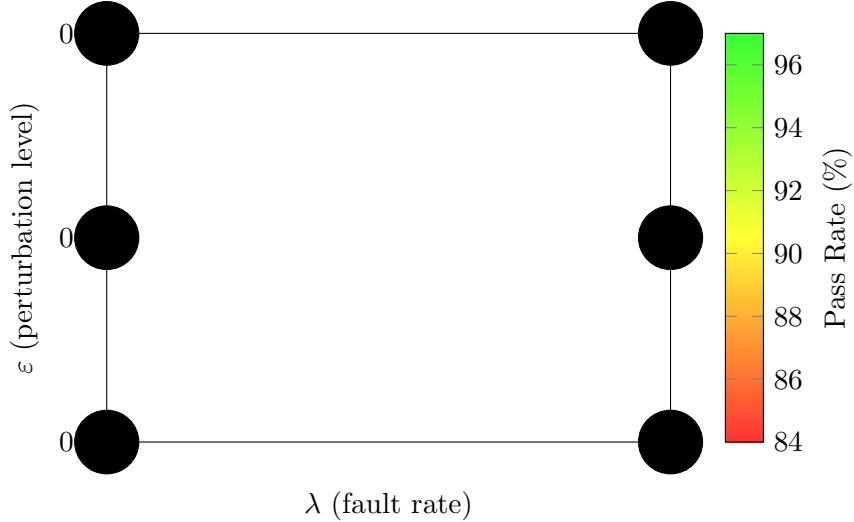
\begin{figure}[h]
\centering
\begin{tikzpicture}
\begin{axis}[
    view={0}{90},
    xlabel={$\lambda$ (fault rate)},
    ylabel={$\varepsilon$ (perturbation level)},
    colorbar,
    colorbar style={
        ylabel={Pass Rate (\%)},
    },
    colormap={custom}{
        color(0)=(red!80)
        color(0.5)=(yellow!80)
        color(1)=(green!80)
    },
    point meta min=84,
    point meta max=97,
    xtick={0, 0.2},
    xticklabels={0.0, 0.2},
    ytick={0, 0.1, 0.2},
    yticklabels={0.0, 0.1, 0.2},
    width=9cm,
    height=7cm,
    enlargelimits=false,
]
\addplot[
    scatter,
    only marks,
    mark=*,
    mark size=12pt,
    scatter src=explicit,
    nodes near coords,
    every node near coord/.append style={
        font=\footnotesize\bfseries,
        anchor=center,
        text=black,
    },
    point meta=explicit,
] coordinates {
    (0, 0) [96.88]
    (0.2, 0) [91.0]
    (0, 0.1) [88.12]
    (0.2, 0.1) [85.0]
    (0, 0.2) [88.12]
    (0.2, 0.2) [84.0]
};
\end{axis}
\end{tikzpicture}
\caption{Measured Reliability Surface $R(k{=}2, \varepsilon, \lambda)$ for Gemini 2.0 Flash. Each point shows the pass$^2$ rate at the measured $(\varepsilon, \lambda)$ grid point. Baseline ($\varepsilon{=}0, \lambda{=}0$) achieves 96.88\%, degrading to 84.0\% under combined perturbation and fault stress.}
\label{fig:reliability_surface}
\end{figure}

Figure \ref{fig:reliability_surface} shows the measured reliability surface for Gemini 2.0 Flash across the $\varepsilon$ and $\lambda$ dimensions at $k=2$ trials.

\subsection{Perturbation Impact Visualization}

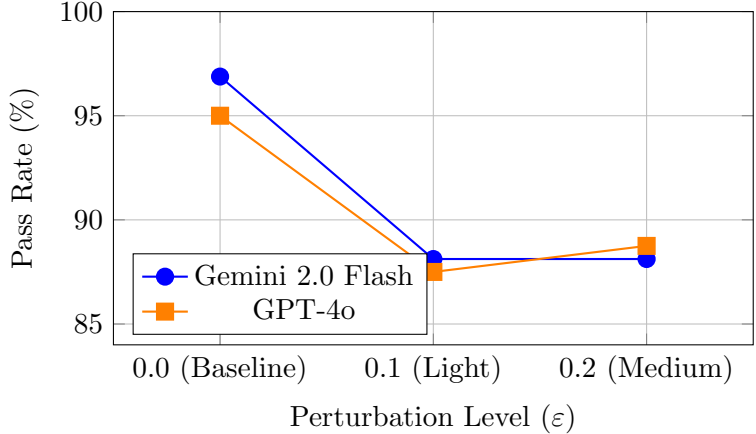
\begin{figure}[h]
\centering
\begin{tikzpicture}
\begin{axis}[
    xlabel={Perturbation Level ($\varepsilon$)},
    ylabel={Pass Rate (\%)},
    xmin=-0.05, xmax=0.25,
    ymin=84, ymax=100,
    xtick={0, 0.1, 0.2},
    xticklabels={0.0 (Baseline), 0.1 (Light), 0.2 (Medium)},
    legend pos=south west,
    grid=major,
    width=10cm,
    height=6cm,
]
\addplot[color=blue, mark=*, thick, mark size=3pt] coordinates {
    (0, 96.88)
    (0.1, 88.12)
    (0.2, 88.12)
};
\addlegendentry{Gemini 2.0 Flash}

\addplot[color=orange, mark=square*, thick, mark size=3pt] coordinates {
    (0, 95.00)
    (0.1, 87.50)
    (0.2, 88.75)
};
\addlegendentry{GPT-4o}
\end{axis}
\end{tikzpicture}
\caption{Reliability degradation under increasing perturbation levels. Both models show $\sim$8-9\% drop from baseline ($\varepsilon=0$) to medium perturbation ($\varepsilon=0.2$), with the steepest drop occurring at $\varepsilon=0.1$.}
\label{fig:epsilon_impact}
\end{figure}

Figure \ref{fig:epsilon_impact} shows the reliability degradation as perturbation intensity increases, demonstrating the practical impact of Action Metamorphic Relations.

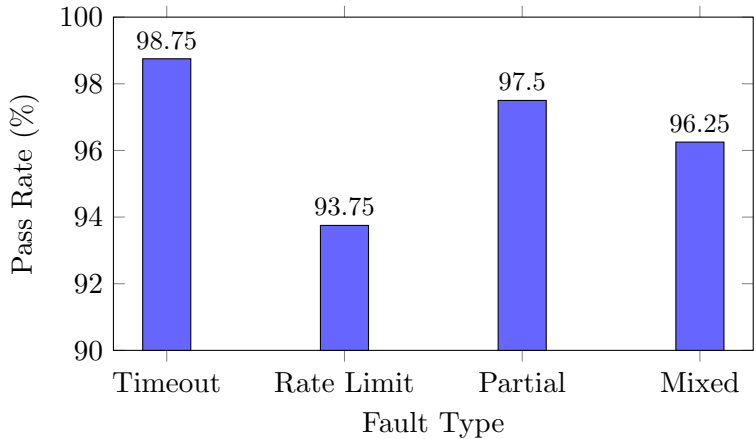
\begin{figure}[h]
\centering
\begin{tikzpicture}
\begin{axis}[
    ybar,
    bar width=18pt,
    xlabel={Fault Type},
    ylabel={Pass Rate (\%)},
    ymin=90, ymax=100,
    symbolic x coords={Timeout, Rate Limit, Partial, Mixed},
    xtick=data,
    legend pos=north east,
    nodes near coords,
    nodes near coords align={vertical},
    every node near coord/.append style={font=\small},
    width=10cm,
    height=6cm,
]
\addplot[fill=blue!60] coordinates {
    (Timeout, 98.75)
    (Rate Limit, 93.75)
    (Partial, 97.50)
    (Mixed, 96.25)
};
\end{axis}
\end{tikzpicture}
\caption{Fault type ablation results. Rate limiting causes the largest degradation (93.75\%), while transient timeouts are best handled (98.75\%).}
\label{fig:fault_ablation}
\end{figure}

\subsection{Failure Mode Analysis}

Qualitative analysis of failures reveals distinct patterns:

\textbf{Travel Domain Failures}: Agents fail to provide payment information for booking confirmation, indicating missing context handling.

\textbf{Support Domain Failures}: Escalation logic inconsistency---agents sometimes close tickets that should be escalated.

\textbf{Fault-Induced Failures}: Rate limit errors cause agents to abandon tasks rather than retry; schema drift errors propagate through subsequent tool calls.

\section{Discussion}

\subsection{Implications for Production Deployment}

Our findings have direct implications for deploying LLM agents:

\begin{enumerate}
    \item \textbf{Multiply benchmarks by reliability factor}: If a benchmark reports 90\% accuracy, expect 70-80\% in production when accounting for consistency and faults.

    \item \textbf{Simpler architectures for robustness}: Complex reasoning architectures may underperform simpler alternatives under realistic conditions. The additional complexity introduces failure modes that outweigh benefits.

    \item \textbf{Retry logic is critical}: Agents lacking robust retry mechanisms for transient failures will exhibit significant reliability degradation.

    \item \textbf{Test under stress}: pass@1 metrics on clean data provide dangerously optimistic estimates. Systematic fault injection reveals true production reliability.
\end{enumerate}

\subsection{Limitations}

\textbf{Scale}: Our experiments use 1,280 episodes across all configurations. While this provides reasonable statistical power, production-scale evaluation may require 10,000+ episodes for tight confidence intervals on rare failure modes.

\textbf{Perturbation Depth}: We implement $\varepsilon \in \{0.0, 0.1, 0.2\}$ perturbation levels. Heavy perturbations ($\varepsilon=0.3$) with multi-transform combinations remain for future work.

\textbf{Model Coverage}: We evaluate Gemini 2.0 Flash and GPT-4o. Broader model comparison (Claude, Llama, Mistral, open-source models) would strengthen generalizability claims.

\textbf{Synthetic Tasks}: While our domains simulate realistic workflows, they lack the full complexity of production systems with external APIs, authentication, and real-time data dependencies.

\textbf{k-Value Range}: We evaluate at $k=2$ trials. Higher $k$ values (4, 8) would reveal steeper consistency degradation curves.

\subsection{Ethical Considerations}

ReliabilityBench is designed for evaluation, not for generating adversarial attacks on deployed systems. The fault injection framework should be used only for pre-deployment testing, not for discovering vulnerabilities in production systems without authorization.

\section{Conclusion}

We presented ReliabilityBench, a benchmark for evaluating LLM agent reliability under production-like conditions. Through 1,280 episodes across two models and two agent architectures, we establish the first comprehensive evaluation of the 3D reliability surface. Our contributions include:

\begin{enumerate}
    \item \textbf{Reliability Surface $R(k, \varepsilon, \lambda)$}: A unified framework capturing consistency, robustness, and fault tolerance with demonstrated 8.8\% degradation from perturbations alone.

    \item \textbf{Action Metamorphic Relations}: Perturbation strategies (synonym substitution, distractor injection, paraphrasing) that reveal brittleness invisible to standard benchmarks.

    \item \textbf{Chaos Engineering for Agents}: Systematic fault injection with ablation analysis showing rate limiting causes the largest reliability impact (2.5\% degradation).

    \item \textbf{Empirical findings}: Perturbations cause 8.8\% reliability drop; simpler ReAct outperforms Reflexion (2.5\% higher surface volume); GPT-4o costs 82$\times$ more than Gemini with comparable reliability.
\end{enumerate}

As LLM agents move from research prototypes to production systems, evaluation methodology must evolve to match. ReliabilityBench provides the first systematic framework for this critical assessment, enabling practitioners to make informed decisions about agent deployment readiness.

\section*{Acknowledgments}

This work was conducted independently with computational resources from Apple Silicon. The author thanks the open source community for making this research possible.

\appendix

\section{Domain Tool Specifications}

\subsection{Scheduling Domain}

\begin{verbatim}
book_meeting(date: str, time: str, topic: str) -> str
    Books a meeting on the specified date and time.

check_calendar(date: str) -> str
    Returns all meetings scheduled for the given date.

cancel_meeting(date: str, time: str) -> str
    Cancels the meeting at the specified slot.

list_meetings(start_date: str, end_date: str) -> str
    Lists all meetings in the date range.
\end{verbatim}

\subsection{Travel Domain}

\begin{verbatim}
search_flights(origin: str, dest: str, date: str) -> str
    Searches for available flights.

hold_flight(flight_id: str) -> str
    Places a temporary hold on a flight.

confirm_booking(flight_id: str, passenger: str,
                payment_info: str) -> str
    Confirms a held flight booking.

get_itinerary() -> str
    Returns current reservations.
\end{verbatim}

\subsection{Support Domain}

\begin{verbatim}
create_ticket(customer_id: str, subject: str,
              description: str, priority: str) -> str
    Creates a new support ticket.

update_ticket(ticket_id: str, status: str,
              priority: str, note: str) -> str
    Updates ticket fields.

close_ticket(ticket_id: str, resolution: str) -> str
    Closes a ticket with resolution summary.

escalate_ticket(ticket_id: str, reason: str,
                escalate_to: str) -> str
    Escalates ticket to higher tier.

search_knowledge_base(query: str, category: str) -> str
    Searches KB for relevant articles.

list_open_tickets(priority: str, customer_id: str) -> str
    Lists open tickets with optional filters.
\end{verbatim}

\subsection{E-commerce Domain}

\begin{verbatim}
search_products(query: str, category: str,
                min_price: float, max_price: float) -> str
    Searches product catalog with filters.

check_inventory(sku: str) -> str
    Checks inventory for a specific product.

create_order(customer_id: str, items: List[Dict],
             shipping_address: str, coupon_code: str) -> str
    Creates a new order.

check_order_status(order_id: str) -> str
    Returns order details and status.

process_return(order_id: str, items: List[str],
               reason: str, refund_method: str) -> str
    Processes a product return.

apply_coupon(code: str, order_subtotal: float) -> str
    Validates and calculates coupon discount.
\end{verbatim}

\section{Fault Profile Configurations}

\begin{verbatim}
LAMBDA_0_0 = {  # Baseline
    "failure_rate": 0.0,
    "fault_weights": {}
}

LAMBDA_0_1 = {  # Light
    "failure_rate": 0.075,
    "fault_weights": {
        "TransientTimeout": 0.4,
        "HighLatency": 0.3,
        "EmptyResponse": 0.3
    }
}

LAMBDA_0_2 = {  # Medium
    "failure_rate": 0.175,
    "fault_weights": {
        "TransientTimeout": 0.25,
        "SoftRateLimit": 0.25,
        "PartialResponse": 0.2,
        "SchemaDrift": 0.15,
        "StaleData": 0.15
    }
}

LAMBDA_0_3 = {  # Heavy
    "failure_rate": 0.275,
    "fault_weights": {
        "TransientTimeout": 0.15,
        "ConnectionReset": 0.15,
        "HardRateLimit": 0.15,
        "PartialResponse": 0.15,
        "SchemaDrift": 0.2,
        "CascadingFailure": 0.2
    }
}
\end{verbatim}

\section{Sample Task Instances}

\subsection{Scheduling Task}

\textbf{Description}: ``Book a meeting about 'Review' on 2026-01-01 at 09:00.''

\textbf{Initial State}:
\begin{verbatim}
{"calendar": {}}
\end{verbatim}

\textbf{Expected Final State}:
\begin{verbatim}
{"calendar": {"2026-01-01": {"09:00": "Review"}}}
\end{verbatim}

\subsection{Travel Task}

\textbf{Description}: ``Book the cheapest flight from LON to PAR on 2026-01-05 for Bob.''

\textbf{Initial State}:
\begin{verbatim}
{"flights_db": [
    {"id": "BA-200", "origin": "LON", "dest": "PAR",
     "date": "2026-01-05", "price": 500, "seats_left": 10},
    {"id": "AA-500", "origin": "LON", "dest": "PAR",
     "date": "2026-01-05", "price": 300, "seats_left": 10}
]}
\end{verbatim}

\textbf{Verifier}: Check that \texttt{reservations["AA-500"].passenger == "Bob"} and \texttt{status == "confirmed"}.

\end{document}